\documentclass[10pt,twocolumn]{article}

\usepackage[margin=0.72in]{geometry}
\usepackage{amsmath}
\usepackage{booktabs}
\usepackage{graphicx}
\usepackage{enumitem}
\usepackage{microtype}
\usepackage{xcolor}
\usepackage{hyperref}
\usepackage{url}

\hypersetup{colorlinks=true,linkcolor=blue,citecolor=blue,urlcolor=blue}
\setlist{nosep,leftmargin=*}
\newcommand{\system}{DLFP}
\newcommand{\good}{\textcolor{teal!70!black}{\textbf{verified}}}

\title{Decode-Latency Feedback Prefill:\\
A Model-Free Controller and Its Generalization Limits}

\author{
Gaurav Agarwal\\
\texttt{gaurav.iit219@gmail.com}
\and
Ashish Garg, Ph.D.\\
\texttt{ashish.philly@gmail.com}
\and
Isha Singhal\\
\texttt{ishasinghal22@gmail.com}
}
\date{}

\begin{document}
\maketitle

\begin{abstract}
Concurrent autoregressive inference creates a fundamental interference problem:
prefilling a newly arrived long prompt can delay tokens for requests that are
already decoding. Fixed prefill chunks reduce this interference, but the best
chunk size depends on the model, hardware, load, and latency objective. We
introduce \emph{Decode-Latency Feedback Prefill} (\system), a model-free
controller that changes only prefill work that overlaps active decodes. After a
guarded scheduling cycle, \system{} uses the observed interval as proportional
feedback to resize the next prefill chunk; isolated prefills remain unrestricted.

We implement \system{} in vLLM and evaluate it with open-loop Poisson arrivals,
exact token accounting, raw request traces, and NVIDIA telemetry. On Qwen3-0.6B
in BF16 on one A100 80~GB GPU, three paired 100-request trials reduce P99
inter-token latency by 24.8\%, 30.1\%, and 28.2\% (mean 27.7\%, paired 95\%
confidence interval 21.0--34.3\%) with exact output agreement, no failures, and
unchanged SLO compliance. The benefit is not free: mean P99 time to first token
increases 34.8\% while remaining inside the declared SLO.

Crucially, the mechanism does not generalize to Qwen3-8B, Qwen3-32B, or a
two-GPU tensor-parallel configuration. We trace the failure to an asynchronous
scheduler-call interval that is only a proxy for completed GPU iteration time.
This negative result defines the boundary of the contribution and motivates a
completion-timed controller for concurrent CPU and on-device inference. We do
not claim mobile-device performance; the present work is a reproducible
proof-of-concept and generalization study.
\end{abstract}

\section{Introduction}

Large language model inference alternates between two operationally different
phases. \emph{Prefill} processes all prompt tokens and exposes substantial
parallelism. \emph{Decode} generates one token per active sequence per
iteration and is sensitive to any work added to the critical path. Continuous
batching improves utilization, but a long prompt admitted beside active decodes
can turn one iteration into a visible pause for every interactive user in that
batch \cite{orca,vllm,sarathi}.

This problem is increasingly relevant outside datacenters. A laptop assistant
may generate an answer while indexing a document; a phone may run an interactive
agent beside a background summarizer. Small open-weight models make such
concurrency plausible, but edge devices add strict responsiveness, power, and
thermal constraints. A useful scheduler should therefore protect decode latency
without relying on an expensive per-device latency model.

Fixed-size chunked prefill is an effective baseline \cite{sarathi}. Its tuning,
however, is fragile. Large chunks maximize prefill throughput but create decode
stalls. Small chunks add repeated scheduling and kernel overhead, and can move
latency from inter-token delay into time to first token (TTFT). Recent systems
use predictors, deadline state, or offline calibration to choose chunks
\cite{lprs,sloweave}. We ask a narrower question: \emph{can recent observed
latency directly close the loop?}

We make three contributions:

\begin{enumerate}
  \item We design and implement \system{}, a proportional, model-free controller
  that caps only prefills overlapping active decodes and resets when interference
  disappears.
  \item We provide a controlled end-to-end evaluation with deterministic token
  inputs, open-loop Poisson arrivals, three paired trials, confidence intervals,
  exact-output checks, and retained raw telemetry.
  \item We report both the positive small-model result and failed 8B, 32B, and
  two-GPU transfer experiments. The failures identify scheduler-call cadence as
  a non-portable feedback signal under asynchronous execution.
\end{enumerate}

The resulting claim is intentionally bounded. \system{} is \good{} for the
Qwen3-0.6B single-GPU operating point tested here. It is not yet a generally
validated serving policy, and it has not been evaluated on a phone or CPU-only
laptop.

\section{Background and Motivation}

\subsection{Continuous batching and prefill interference}

Let $D_t$ be the set of requests decoding at scheduler iteration $t$, and let
$p_t$ be the number of prompt tokens admitted for prefill. A simplified mixed
iteration time is
\begin{equation}
  L_t \approx L_{\mathrm{decode}}(D_t) +
  L_{\mathrm{prefill}}(p_t,D_t) + L_{\mathrm{overhead}}.
\end{equation}
Every request in $D_t$ observes approximately $L_t$ before receiving its next
token. A large $p_t$ can therefore create a correlated latency spike across all
active decoders even when the admission queue is empty.

PagedAttention reduces KV-cache fragmentation and enables high-concurrency
serving \cite{vllm}. Iteration-level scheduling and continuous batching, as in
Orca \cite{orca}, improve utilization. Sarathi-Serve shows that chunked prefill
can form decode-maximal batches and tame the throughput--latency tradeoff
\cite{sarathi}. Prefill/decode disaggregation instead separates the phases
across resources \cite{distserve}, which is attractive at datacenter scale but
not always available on a single device.

\subsection{Why feedback control?}

An analytical or learned latency model must cover model size, attention shapes,
batch composition, kernels, device generation, clock state, and competing work.
For heterogeneous laptops and phones, maintaining such a model can be more
costly than the controller itself. Feedback control offers a tempting
alternative: observe a latency consequence, adjust the prefill budget, and
repeat. The central risk is measurement fidelity. If the observation is not the
actual completed device latency, the controller can confidently move in the
wrong direction. Our generalization study makes this risk concrete.

\section{Design}

\subsection{Control policy}

\system{} activates only when at least one request is already decoding and a
prefill would share the next scheduling cycle. Let $c_t$ be the current prefill
cap in tokens, $T$ the target mixed-iteration latency, and $\hat{L}_t$ the
observed interval following a guarded cycle. The next cap is
\begin{equation}
 c_{t+1} = \operatorname{clamp}\left(
 \operatorname{floor}_{128}\left(c_t \frac{T}{\hat{L}_t}\right),
 c_{\min}, c_{\max}\right).
\end{equation}
If the observation exceeds the target, the next chunk shrinks; if it is below
target, the chunk grows. Quantization to 128-token increments avoids unstable
fine-grained changes. When no decoder is active, the controller resets to
$c_{\mathrm{initial}}$ and does not cap isolated prefill.

The evaluated parameters are $T=80$~ms,
$c_{\mathrm{initial}}=3072$, $c_{\min}=1024$, and $c_{\max}=6144$. A retained
observability run recorded a live update from 3072 to 4224 tokens after a
57.51~ms observation, confirming that the controller did not remain a static
cap.

\subsection{Scheduler integration}

The implementation modifies the vLLM V1 scheduler. It first detects whether a
running request has completed its prompt. For both running chunked-prefill
requests and newly admitted requests, it limits prompt tokens to the current cap
only if a decoder is active. Decode scheduling, model weights, BF16 precision,
attention backend, sampling, KV layout, and isolated-prefill scheduling are
unchanged.

The prototype uses elapsed scheduler-call cadence as $\hat{L}_t$. This choice is
cheap and appeared correlated with iteration duration on the initial small-model
configuration. Under asynchronous scheduling, however, a later scheduler call
can occur before or after different amounts of queued GPU work. Section~\ref{sec:generalization}
shows why this approximation is the principal design limitation.

\section{Experimental Methodology}

\subsection{Hardware and software}

The main experiment uses one NVIDIA A100 80~GB GPU. The server runs vLLM 0.17.1,
PyTorch 2.10, CUDA 12.8 user-space libraries, FlashAttention-2, CUDA graphs, and
asynchronous scheduling. The public Qwen3-0.6B checkpoint \cite{qwen3} is served
in BF16 with maximum model length 18,000, maximum 64 sequences, and an
8,192-token global batch budget. Prefix caching is disabled. Startup and graph
compilation are excluded from measurement.

Generalization experiments use Qwen3-8B and Qwen3-32B in BF16 across two A100
80~GB GPUs connected through a PCIe host bridge without NVLink. We also test
Qwen3-0.6B with tensor parallel size two.

\subsection{Workloads and SLOs}

We construct exact-token workloads for Chat (1,024 input/128 output), RAG
(8,192/256), Agent (16,384/256), and Decode-heavy (256/1,024). Requests arrive
open-loop according to independent exponential inter-arrival times. The primary
comparison uses nonmatching Agent prompts, offered load 1.0 request/s, 100
measured requests after three warmups, and seeds 801, 802, and 803.

The Agent SLO is TTFT $\leq 5$~s, mean inter-token latency (ITL) $\leq 35$~ms,
and end-to-end latency $\leq 14$~s. SLO-compliant goodput counts only requests
meeting all three conditions. We record each token ID and timestamp, HTTP and
engine failures, token usage, queue depth, KV occupancy, GPU utilization,
memory, power, and sampled energy.

\subsection{Comparison discipline}

Within each paired seed, baseline and candidate use identical model, prompts,
arrival schedule, output length, precision, hardware, and measurement code.
Every candidate must reconcile request counts and prompt/output tokens, introduce
no failures, and match greedy output IDs where deterministic agreement is
expected. We report Student's $t$ 95\% confidence intervals over three paired
trial-level values ($df=2$). P99 results from small reconnaissance runs are not
used as final claims.

\section{Results}

\subsection{Primary latency result}

Table~\ref{tab:main} reports the held-out comparison. All 600 baseline and
candidate requests complete successfully, all workload fingerprints match, and
all 76,800 paired generated token IDs agree exactly.

\begin{table}[t]
\centering
\caption{Qwen3-0.6B single-GPU Agent results. Lower P99 ITL is better.}
\label{tab:main}
\small
\begin{tabular}{rrrrr}
\toprule
Seed & Baseline & \system{} & Reduction & SLO pass \\
 & \multicolumn{2}{c}{P99 ITL (ms)} & & \\
\midrule
801 & 122.22 & 91.95 & 24.8\% & 100/100 \\
802 & 187.07 & 130.85 & 30.1\% & 100/100 \\
803 & 200.36 & 143.84 & 28.2\% & 100/100 \\
\midrule
Mean & --- & --- & 27.7\% & 300/300 \\
\bottomrule
\end{tabular}
\end{table}

Mean paired P99 ITL reduction is 27.7\%, with a 95\% confidence interval of
21.0--34.3\%. Every trial independently exceeds the predeclared 10\% threshold.
Mean goodput changes by $-0.10$\% (95\% CI $-0.50$ to $+0.29$\%), and energy per
output token changes by $+0.05$\% (95\% CI $-1.55$ to $+1.66$\%); neither is a
claim.

\begin{figure}[t]
\centering
\includegraphics[width=\columnwidth]{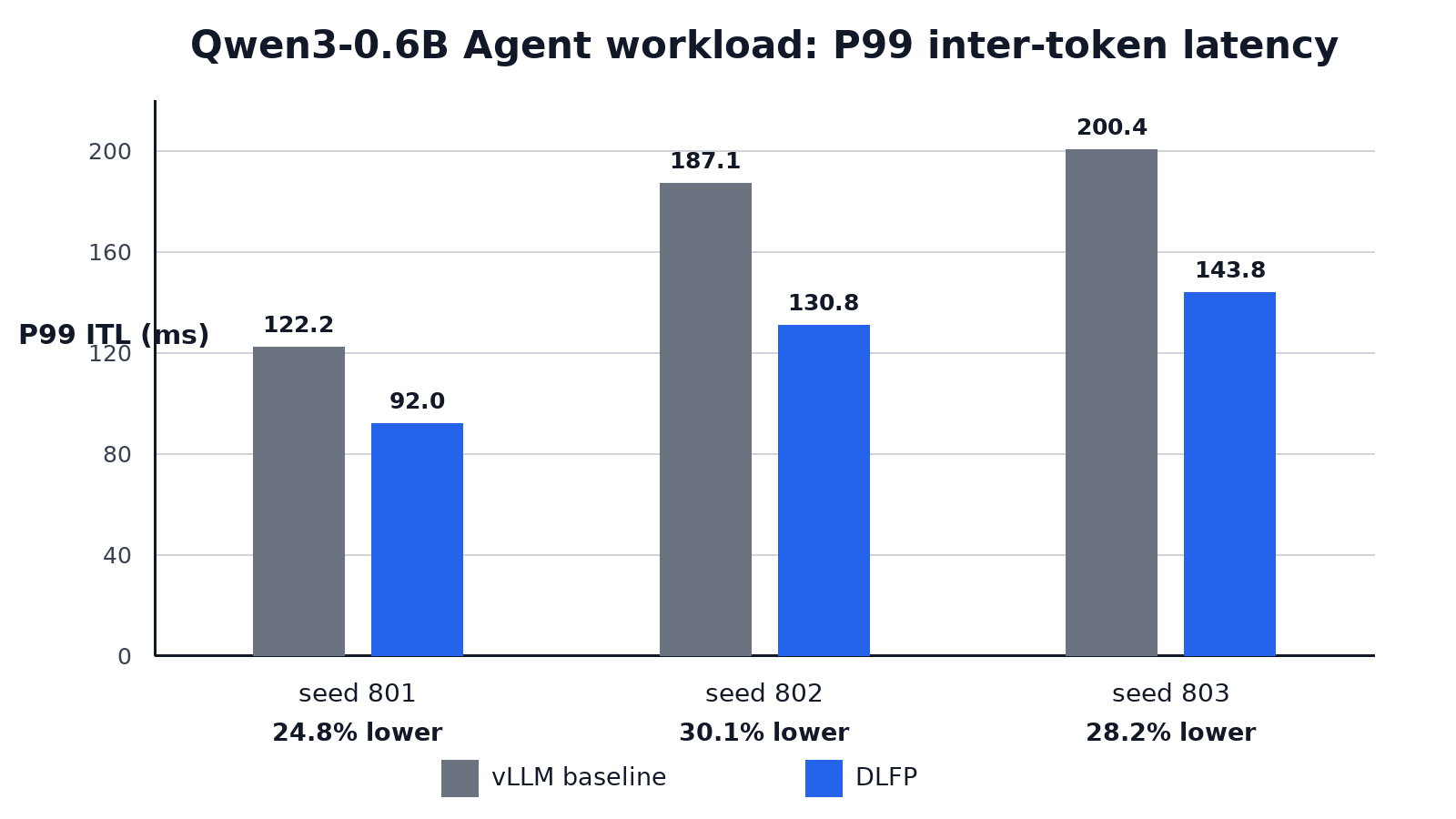}
\caption{Held-out P99 ITL for the three paired trials. Each trial independently
clears the 10\% reduction gate.}
\label{fig:main}
\end{figure}

\subsection{Tradeoff}

\system{} prioritizes decode smoothness. Mean P99 TTFT increases 34.8\% (95\% CI
3.6--65.9\%), while mean P99 end-to-end latency increases 7.8\% (95\% CI
$-3.5$--19.2\%). Both remain inside the declared Agent SLO for all final
requests. This tradeoff rules out unconditional deployment: \system{} is useful
only when interactive token cadence is more important than minimum TTFT.

\subsection{Controls}

Automatic prefix caching substantially improves matching-prefix RAG TTFT but
returns to baseline on nonmatching prompts; it validates the harness but is an
established technique. Fixed prefill caps can reduce ITL, but a 2,048-token cap
reduces aggregate SLO compliance to 287/300. A 4,096-token cap fails the repeated
10\% gate, while a 3,072-token cap loses SLO passes and is not a new mechanism.

\section{Generalization and Negative Results}
\label{sec:generalization}

\subsection{Two-GPU Qwen3-0.6B}

With tensor parallel size two, paired P99 ITL reductions are 5.3\%, 8.2\%, and
24.7\%. The 12.7\% mean has a confidence interval crossing zero. Candidate SLO
passes fall from 110/300 to 101/300, P99 TTFT rises 44.1\%, and energy per output
token rises 4.3\%. This is not a win.

\subsection{Qwen3-8B}

At 0.10 request/s, baseline P99 ITL is 19~ms; \system{} raises it to 281~ms and
reduces SLO passes from 10/20 to 7/20. There is little interference to remove,
and fragmented prefills create additional tail events. At 0.25 request/s,
\system{} reduces P99 ITL from 1.763~s to 707~ms, but P99 TTFT rises from 12.33
to 14.81~s and SLO passes fall from 2/20 to 0/20. The controller moves latency
rather than restoring useful goodput.

\subsection{Qwen3-32B}

At 0.05 request/s on two GPUs, baseline/\system{} P99 values are 18.07/21.64~s
TTFT, 314/670~ms ITL, and 45.61/45.86~s end-to-end. A model-specific SLO yields
5/12 passes in both configurations, and energy per output token regresses 4.7\%.

A completion-aware revision updates the controller when the scheduler output
returns and bounds each control step. It also fails. Under synchronous
scheduling, baseline P99 ITL is 41.1~ms versus 194.8~ms for the candidate, with
SLO passes falling from 5/12 to 4/12. The evidence rejects simple retuning as a
solution.

\subsection{Root cause}

The small-model success depended on an accidental correlation between
scheduler-call cadence and completed GPU iteration time. Larger kernels,
asynchronous queueing, and tensor-parallel synchronization break that
correlation. The proportional update can then expand the cap after observing a
short host interval even when device work is still outstanding. More generally,
splitting one large stall into several moderate stalls can increase the fraction
of token gaps that reach the global P99.

\section{Implications for Concurrent Edge Inference}

The motivating use case remains promising, but the current implementation is
not a mobile result. A single interactive phone session has no overlapping
prefill and decode, so \system{} cannot help. The opportunity appears when
multiple local tasks overlap: foreground generation plus document ingestion,
multiple agent branches, or background summarization.

CPU-only laptops are a particularly useful next target. A synchronous CPU
runtime can expose actual completion time directly, avoiding the asynchronous
GPU proxy that caused our failures. Mobile GPU and NPU runtimes require device
events or runtime completion callbacks, plus thermal and energy measurements.
Candidate runtimes include llama.cpp \cite{llamacpp}, MLC LLM \cite{mlcllm}, and
ExecuTorch \cite{executorch}.

A publishable edge evaluation should include several 0.5B--3B quantized models,
multiple phones and CPU laptops, sustained thermal runs, foreground/background
priority, and strong fixed-chunk controls. The acceptance criterion should
combine P99 ITL, TTFT, SLO goodput, energy per token, and quality. Until those
experiments exist, our result should be described as evidence motivating
completion-timed adaptive prefill for edge concurrency, not as proof of edge
speedup.

\section{Related Work}

Iteration-level scheduling and continuous batching were developed in systems
such as Orca \cite{orca}; vLLM's PagedAttention improves KV-cache memory
management and serving throughput \cite{vllm}. Sarathi-Serve introduces
decode-maximal batching and chunked prefill \cite{sarathi}. DistServe separates
prefill and decode to optimize goodput \cite{distserve}, while Splitwise studies
phase separation on heterogeneous resources \cite{splitwise}.

Recent adaptive chunking approaches select prefill work using latency models,
fairness objectives, or decode deadlines. LPRS uses an offline feature-based
latency predictor and active prefill control \cite{lprs}. SLOWeave chooses chunks
against request deadlines using a monotone cost model \cite{sloweave}. \system{}
differs in using a direct proportional update without an offline model or
per-request deadline state. Adaptive chunk sizing itself is therefore prior art;
our claim is the specific model-free controller and its measured behavior, not
the broad idea of adaptive prefill.

On-device runtimes focus primarily on compilation, kernels, quantization, and
portable deployment \cite{mlcllm,llamacpp,executorch}. Our proposed edge
direction is complementary: schedule concurrent request phases after the model
already runs efficiently on the device.

\section{Limitations and Threats to Validity}

The verified positive result covers one small model, one server GPU, one primary
workload and rate, three trials, and 100 requests per trial. GPU clocks were not
locked. NVML power is sampled. The controller's observation does not measure
device completion, and the failed larger-model experiments show that the result
must not be generalized by model size or device count.

The workloads use deterministic synthetic token sequences to ensure exact
comparability; production prompts may change attention shapes, output entropy,
and stopping behavior. Temperature zero enables a strong exact-output check but
does not replace a semantic quality evaluation for quantized or speculative
systems. P99 estimates from 100 requests remain less precise than a large
production trace, although each request contains 255 ITL observations.

Finally, a bounded literature search cannot establish worldwide novelty or
patentability. We found close work on fixed and adaptive chunking. We therefore
describe \system{} as a differentiated research prototype and report its
generalization failures prominently.

\section{Reproducibility and Ethics}

The artifact contains the scheduler patch, environment manifest, model hashes,
benchmark generator, exact commands, raw JSON request records, telemetry,
summary scripts, and negative results. No proprietary model or private user data
is used; all models are public Qwen checkpoints and prompts are synthetic. No
driver modification is required. The benchmark refuses to run when unrelated
GPU processes are present.

The system does not change model outputs intentionally. All final positive-run
tokens match exactly. The primary ethical risk is overclaiming: presenting an
A100 result as a phone result could mislead deployment decisions. We explicitly
avoid that claim.

\section{Conclusion}

\system{} demonstrates that lightweight feedback can materially improve decode
smoothness at one small-model operating point: mean P99 ITL falls 27.7\% with
exact outputs, no failures, and unchanged SLO compliance. Equally important,
the mechanism fails on larger models and multi-GPU execution because its host
scheduler interval is not a reliable device-completion signal. The next step is
not broader deployment of the current patch; it is a completion-timed controller
evaluated on real concurrent CPU and mobile workloads. Reporting both outcomes
turns a promising isolated speedup into a useful systems result with a clear
research path.

\bibliographystyle{plain}
\bibliography{references}

\end{document}